\documentclass[conference]{IEEEtran}
\IEEEoverridecommandlockouts

\usepackage{cite}
\usepackage{times}
\usepackage{latexsym}
\usepackage[T1]{fontenc}
\usepackage[utf8]{inputenc}
\usepackage{microtype}
\usepackage{inconsolata}

\usepackage{amsmath,amssymb,mathtools}
\usepackage{booktabs,multirow,makecell,array}
\usepackage[table]{xcolor}
\usepackage{graphicx}
\usepackage{xspace}
\usepackage{enumitem}
\usepackage[most]{tcolorbox}
\usepackage{pifont}
\usepackage{placeins}
\usepackage{float}
\usepackage{subfigure}
\usepackage{url}
\usepackage{pgfplots}
\pgfplotsset{compat=1.18}
\usepackage{colortbl}
\usepackage{algorithm}
\usepackage{algorithmic}
\usepackage{tabularx}
\usepackage{xcolor}
\usepackage[colorlinks=true, urlcolor=blue, linkcolor=black, citecolor=black]{hyperref}

\newcolumntype{C}{>{\centering\arraybackslash}X}

\setlist[itemize]{leftmargin=*, itemsep=1.5pt, topsep=2pt}

\newcommand{\ourmethod}{{DiFA}\xspace}
\newcommand{\sem}{\mathrm{sem}}
\newcommand{\form}{\mathrm{form}}

\definecolor{desaBlue}{HTML}{1F5FA8}
\definecolor{desaLightBlue}{HTML}{EAF2FF}
\definecolor{desaOrange}{HTML}{D7651D}
\definecolor{desaLightOrange}{HTML}{FFF3E8}
\definecolor{desaGreen}{HTML}{2F6B3A}
\definecolor{desaLightGreen}{HTML}{ECF8EF}
\definecolor{desaGray}{HTML}{F5F6F8}

\newtcolorbox{insightbox}[1][]{
  enhanced,
  colback=desaLightBlue,
  colframe=desaBlue,
  boxrule=0.55pt,
  arc=1.2mm,
  left=4pt,right=4pt,top=3pt,bottom=3pt,
  fonttitle=\bfseries,
  title=#1
}

\newtcolorbox{contribbox}{
  enhanced,
  colback=desaGray,
  colframe=black!35,
  boxrule=0.45pt,
  arc=1.2mm,
  left=4pt,right=4pt,top=3pt,bottom=3pt
}

\title{DiFA: Dual Evidence Fusion and Aggregation for Token-Level Text Anomaly Detection}

\author{
\IEEEauthorblockN{
Yanyu Qian$^{*1}$,
Pengcheng Weng$^{*2}$,
Yue Tan$^{*3}$,
Enguang Zuo$^{\dagger 4}$,
Yu Zheng$^{3}$,
and Yixin Liu$^{\dagger 3}$
\thanks{$^{*}$ These authors contributed equally to this work.}
\thanks{$^{\dagger}$ Yixin Liu and Enguang Zuo are the corresponding authors.}
}
\textit{$^1$ Nanyang Technological University, Singapore}
\textit{$^2$ University of Bern, Switzerland} \\
\textit{$^3$ Griffith University, Australia}
\textit{$^4$ Xinjiang University, China} \\
\IEEEauthorblockA{
yanyu003@e.ntu.edu.sg,
pengcheng.weng@students.unibe.ch,
\{yue.tan, yu.zheng, yixin.liu\}@griffith.edu.au,
zeg@xju.edu.cn
}
}

\begin{document}

\maketitle

\begin{abstract}
Text anomaly detection, the task of identifying text instances that deviate from normal language patterns, is crucial for language-driven applications.
However, most existing methods can only perform document-level anomaly detection, making it hard to locate harmful phrases or support targeted prevention. 
Recently, there has been an emerging trend toward token-level text anomaly detection, which aims to address the above limitation by identifying anomalous words or fragments within a document.
Nevertheless, one representative method mainly relies on representation-space distance measurement, neglecting the complementary roles of different anomaly cues in capturing diverse abnormal patterns. 
To bridge the gaps, we propose a \underline{D}ual-ev\underline{i}dence framework with adaptive \underline{F}usion and \underline{A}ggregation (\underline{DiFA}) for token-level anomaly detection. 
\ourmethod derives anomaly scores from form-structural and semantic views to capture visible structural abnormality and contextual inconsistency, respectively, thereby providing complementary evidence for identifying diverse anomalies. 
To combine these two scores with varying numerical scales, \ourmethod incorporates a calibration and fusion mechanism to adaptively balance the two views. 
Moreover, to obtain a discriminative document-level score, a multivariate aggregation method is designed to summarize token-level anomaly scores from multiple perspectives, preventing rare anomalous tokens from being diluted. 
Extensive experiments across various text anomaly detection benchmarks demonstrate that \ourmethod consistently achieves top performance while maintaining strong efficiency, robustness, and interpretability. The code and scripts are available at: \url{https://github.com/qyy11-com/DiFA}.
\end{abstract}


\section{Introduction}

Anomaly detection is a critical task in data mining, with wide applications in fraud detection, cybersecurity, and risk management~\cite{tan2026influence,tan2023taming}. While anomaly detection has been extensively studied on structured data, such as tabular data~\cite{li2026towards}, time series~\cite{wang2025survey}, and graphs~\cite{qiao2025deep,song2024brief,zhao2026fedcigar}, text anomaly detection in the context of natural language processing (NLP) has received relatively limited research attention~\cite{cao2025tad,qian2026dynhd}. 
Taking textual data as input, text anomaly detection aims to identify text instances that deviate from normal language patterns~\cite{cao2025text,cao2025anomaly}. Due to the ubiquity of textual data in real-world applications, text anomaly detection is important for many language-driven applications, such as spam filtering, review moderation, phishing detection, text quality control, and harmful content monitoring~\cite{islam2020deep,chen2025multi}. 

\begin{figure}[t]
    \centering
    \includegraphics[width=\linewidth]{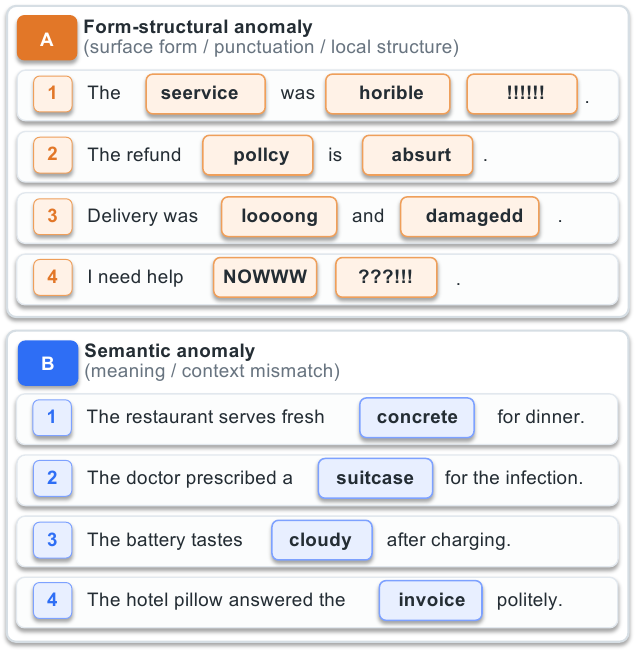}
    \caption{
    Two types of token-level text anomalies.
    }
    \label{fig:intro_examples}
\end{figure}

Modern pretrained language models, particularly transformer-based models, have reshaped text representation learning by producing context-aware embeddings, which serve as a powerful foundation for text anomaly detection~\cite{szoplak2023anomaly,lee2024ced}. Building upon these context-aware representations, recent text anomaly detection methods have shown promising performance in identifying anomalous documents or sentences with lightweight detection modules~\cite{novoa2024explained,wang2024ead}. Nevertheless, most existing methods can only provide \textbf{document-level} anomaly scores for detection, neglecting the need to explain which words or fragments make it suspicious~\cite{cao2026towards}. Such fine-grained predictions can be crucial in many word- or fragment-level scenarios, such as harmful phrase detection and phishing cue identification, where they help practitioners locate the exact suspicious evidence, verify model predictions, and take targeted interventions. 

To enable fine-grained anomaly localization beyond document-level scoring, very recently, Cao et al.~\cite{cao2026towards} proposed the first \textbf{token-level text anomaly detection} approach, TokenCore, which can provide both word-level and document-level anomaly predictions. 
Leveraging representations from pretrained language models (PLMs), TokenCore encodes each token into a contextual embedding and assign anomaly scores by measuring how far each token deviates from normal token patterns. 
As contextual embeddings capture rich semantic and syntactic regularities~\cite{thapa2024review}, tokens that conflict with their surrounding context, or sentiment expressions that are unusual for a review corpus, often become distinguishable in the embedding space, making them easier to identify as token-level anomalies~\cite{cao2026towards}. By directly measuring token deviations in a single representation space, TokenCore demonstrates promising performance in both token- and document-level anomaly detection.

Although TokenCore pioneers token-level text anomaly detection, it relies on simple distance-based scoring within a single representation space, which may overlook the diverse forms of token-level text anomalies~\cite{li2024nlp}. While token-level text anomalies may arise from different linguistic aspects, they can be broadly categorized into two distinct types: form-structural anomalies and semantic anomalies. 
Specifically, \textbf{{form-structural anomalies}} (Block A in Figure~\ref{fig:intro_examples}) are abnormal in their surface form or local structure, exhibiting character corruption, repeated symbols, unusual punctuation, malformed words, or local grammatical violations~\cite{li2024nlp,matthews2024semantics}. 
These signals may be visible in token form, but smoothed out by contextual embeddings optimized for semantic robustness. In contrast, \textbf{semantic anomalies} (Block B in Figure~\ref{fig:intro_examples}) arise when a token is well-formed but inconsistent with the surrounding meaning, domain, or discourse~\cite{xu2023comparative,raghavan2025out}. Such anomalies are usually not reflected in surface form, and instead need to be identified through their incompatibility with the surrounding contextual representations. 
Due to the heterogeneous nature of the two types of anomalies, a token-level detector may require different strategies to capture distinct abnormality patterns; otherwise, relying only on a single representation space with unified scoring can obscure discriminative signals, leading to suboptimal detection performance.

Considering the diverse nature of token-level text anomalies, a natural question arises: \textit{\textbf{Q1 }- how can we design complementary evidence for token-level anomaly detection to capture two types of anomalies?} For semantic anomalies, although PLM-derived representations can encode contextual semantics, directly measuring representation-level deviations may underrepresent token-level semantic conflicts and overlook subtle semantic inconsistencies. For form-structural anomalies, the representations produced by PLMs may smooth out discriminative surface patterns, making it necessary to incorporate non-representational signals that directly model surface-form and local-structural irregularities. 
Even if complementary evidence can be designed for different types of anomalies, a follow-up question remains: \textit{\textbf{Q2} - how can we adaptively fuse these evidence sources to produce indicative anomaly scores for both token-level and document-level prediction?} Due to their heterogeneous origins, different evidence sources may exhibit different scales, distributions, and sensitivities across datasets, making it non-trivial to fuse them into reliable token-level anomaly scores. Moreover, aggregating token-level scores into a document-level prediction is also challenging, since different tokens and their score distributions may contribute unequally to the overall abnormality of a document.

Motivated by the above questions, we propose a novel \textbf{D}ual-ev\textbf{i}dence framework with adaptive \textbf{F}usion and \textbf{A}ggregation (\ourmethod for short) for token-level text anomaly detection. 
Without requiring anomalous examples during training, \ourmethod learns normal token patterns from normal text through two complementary evidence views. 
To answer \textbf{\textit{Q1}}, we design two complementary types of evidence from the PLM-derived representation space and the token-statistic feature space, respectively. More specifically, we introduce 
\textit{form-structural evidence} to capture abnormal surface forms and local structural deviations using compact token descriptors, whereas \textit{semantic evidence} identifies contextual inconsistency through reconstruction and teacher-student discrepancy in the representation space. 
To address \textbf{\textit{Q2}}, we incorporate a calibration-based fusion module to estimate token-level anomaly scores, associated with a multivariate aggregation module for document-level anomaly score prediction. 
To address the scale and distribution mismatch among anomaly scores derived from heterogeneous evidence sources, we introduce a tail-evidence calibration mechanism to transform raw scores into comparable distribution-aware rarity measures, and further design an adaptive fusion module to produce indicative anomaly scores. 
Since document-level anomalies may be triggered by only a few abnormal tokens, \ourmethod further uses multivariate aggregation to model the score distribution of each document, leading to a more flexible measurement of document-level abnormality.
Our contributions are as follows: 
\begin{itemize}
    \item \textbf{Insight.} We identify two major types of token-level text anomalies, i.e., form-structural anomalies and semantic anomalies, and highlight that they require different types of evidence for effective detection.

    \item \textbf{Method.} We propose \ourmethod, a unified framework that constructs complementary evidence from different feature spaces to capture the two types of anomalies. Moreover, we introduce adaptive fusion and aggregation modules to produce token-level and document-level anomaly scores.

    \item \textbf{Experiments.} We conduct extensive experiments on multiple text anomaly detection benchmarks. The results show that \ourmethod achieves superior performance in both token-level and document-level anomaly detection, while maintaining strong efficiency, robustness, and interpretability.
\end{itemize}
\section{Related Work}
\label{sec:related_work}

\subsection{Anomaly Detection}
Anomaly detection aims to identify samples that deviate from the normal data distribution, and has been widely studied in data mining and machine learning~\cite{cao2025anomaly,shen2026raising,liu2026few}.
Classical anomaly detectors rely on different assumptions about abnormality.
Density-based methods such as LOF measure local density deviation~\cite{breunig2000lof}; isolation-based methods such as Isolation Forest exploit the fact that anomalies are easier to isolate by random partitioning; and statistical methods such as ECOD estimate abnormality from distributional tail probabilities~\cite{li2022ecod}.
With the development of deep learning, reconstruction-based and one-class methods have become important paradigms.
Autoencoder-based methods learn to reconstruct normal samples and use reconstruction error as anomaly evidence~\cite{zhou2017anomaly,liu2026rethinking}, while DeepSVDD learns a compact normal region in representation space and detects samples far from this region~\cite{ruff2018deep,pan2026camera}.
Recent neighborhood-based methods, such as LUNAR, further exploit local neighborhood relations to capture complex anomaly structures~\cite{goodge2022lunar,li2026ofa}.

These methods provide general-purpose detection principles and are often used as strong baselines after raw data are transformed into numerical representations.
However, they are mostly designed for generic vector spaces and usually produce instance-level anomaly scores, leaving the modeling of linguistic structure and fine-grained textual evidence to text-specific representations and aggregation mechanisms.

\subsection{Document-Level Text Anomaly Detection}
Text anomaly detection extends general anomaly detection to natural language data, where the goal is to identify textual instances that deviate from normal language patterns~\cite{li2024nlp,cao2025tad,cao2025text,liu2026beyond}.
Existing document-level text anomaly detection methods can be broadly divided into end-to-end and embedding-based approaches.
End-to-end methods directly learn abnormality from raw text or contextual representations.
For example, early neural-network-based methods reconstruct normal documents and detect abnormal texts by reconstruction errors; CVDD learns compact context-vector representations for one-class text detection~\cite{ruff2019self}; DATE uses Transformer-based self-supervised objectives to model normal textual patterns~\cite{manolache2021date}; and FATE introduces deviation learning for few-shot text anomaly detection~\cite{das2023few}.
Embedding-based methods follow a two-step pipeline: a pretrained language model first encodes each document into a dense representation, and then an anomaly detector such as LOF, Isolation Forest, ECOD, AutoEncoder, DeepSVDD, or LUNAR is applied to the resulting embedding space~\cite{li2024nlp,cao2025tad}.
Recent benchmarks show that such embedding-based methods can achieve strong performance because pretrained language models provide rich semantic and syntactic representations~\cite{li2024nlp,cao2025tad}.

Nevertheless, this line of work mainly focuses on document-level decisions: it determines whether a sentence, review, message, or article is anomalous, but does not directly explain which words or fragments are responsible for the prediction.
Moreover, the common embedding--detector pipeline often relies on a single document representation and a fixed detector, which may be sensitive to dataset domains, anomaly types, and the compatibility between embeddings and detectors.

\subsection{Token-Level Text Anomaly Detection}
Token-level text anomaly detection further requires localizing the specific words or fragments that make a document abnormal, while still producing a document-level anomaly decision~\cite{cao2026towards}.
This setting differs from supervised token-level tasks such as spelling correction, grammatical error correction, or named entity recognition, because text anomalies are not restricted to predefined error categories and anomalous tokens are usually unavailable during training.
Instead, the detector is expected to learn normal token patterns from normal documents and identify diverse abnormal tokens at test time, including orthographic corruption, malformed words, unusual punctuation, local structural violations, and semantic inconsistency.
TokenCore is the first representative framework for this setting~\cite{cao2026towards}.
It constructs a memory bank of normal token embeddings and scores each test token by its nearest-neighbor distance to normal token patterns.
Document-level scores are then obtained by aggregating token-level anomaly scores.

Although TokenCore establishes a simple and effective baseline, it also exposes two limitations that motivate our work.
First, relying on a single PLM-derived contextual embedding space may favor semantic or contextual anomalies while smoothing out surface-form and local structural irregularities, since pretrained language models are mainly optimized for semantic representation rather than anomaly sensitivity~\cite{matthews2024semantics,cao2026towards}.
Second, converting sparse token-level evidence into a document-level decision is non-trivial: mean pooling can dilute rare but decisive anomalous tokens, whereas max pooling may overreact to isolated noisy peaks~\cite{cao2026towards}.
In contrast, our method constructs complementary semantic and form-structural evidence for token-level anomaly localization, and further uses multivariate aggregation to transfer token-level evidence into reliable document-level anomaly scores.
\section{Preliminary}
\label{sec:problem_formulation}

\begin{figure*}[t]
    \centering
    \includegraphics[width=1\linewidth]{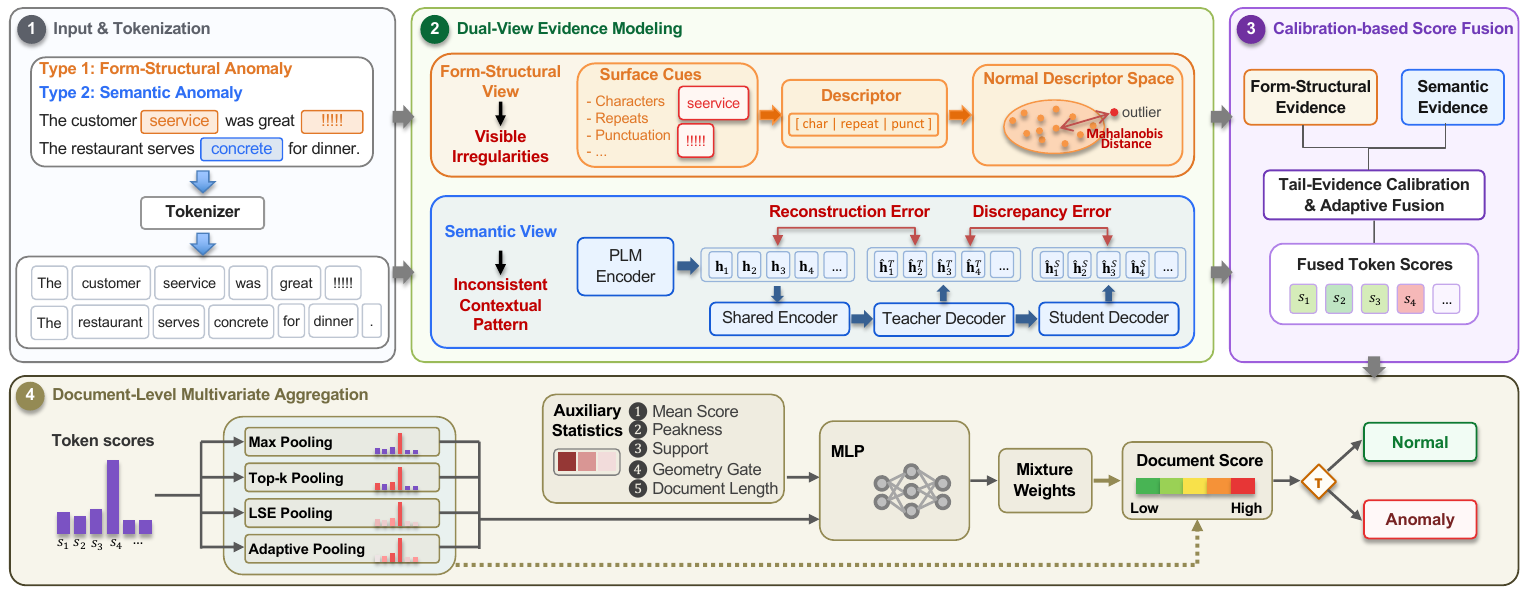}
    \caption{
    Overview of \ourmethod. (1) Two types of text anomalies. (2) A dual-view evidence modeling process that captures visible structural abnormality and contextual inconsistency. 
    The form-structural view measures deviation from an automatically computed descriptor space, while the semantic view measures contextual reconstruction error and teacher--student discrepancy.
    (3) Calibration-based score fusion process that calibrates both views into tail evidence and adaptively fuses them into a single token-level anomaly score. 
    (4) Document-level multivariate aggregation that summarizes token-level anomaly scores into the final document-level anomaly score. 
    }
    \label{fig:method_overview}
\end{figure*}

\noindent\textbf{Document-Level Text Anomaly Detection.}
Traditional text anomaly detection~\cite{li2024nlp} is formulated at the document level.
Let $\mathcal{D}_{n}=\{x_j\}_{j=1}^{N}$ denote a training corpus containing only normal documents.
Given a test document $x$, the goal is to produce a document-level anomaly score $S(x)$, where a larger score indicates that $x$ is more likely to deviate from the normal document distribution.
This setting can determine whether a document is anomalous, but it does not specify which tokens are responsible for the abnormality.

\noindent\textbf{Token-Level Text Anomaly Detection.}
Token-level text anomaly detection~\cite{cao2026towards} extends the traditional setting by requiring fine-grained localization in addition to document-level detection.
Let $x=(w_1,\ldots,w_T)$ denote a document consisting of $T$ word-level tokens, where {token} refers to a natural language word rather than a subword unit.
Given only normal documents during training, the model is expected to output an anomaly score $s_i$ for each token $w_i$, where larger scores indicate stronger token-level abnormality.
The document-level score $S(x)$ is derived from these token-level scores, which also indicate which tokens contribute to the anomaly decision.

\noindent\textbf{Representations for Text Anomaly Detection.} 
As pretrained language models (e.g., BERT) can capture rich semantic features without task-specific supervision, mainstream text anomaly detection methods are usually built upon their representations, followed by lightweight detection models~\cite{li2024nlp}. 
To conduct fine-grained detection, we obtain the word-level contextual representation $\mathbf{h}_i$ by max-pooling the last-layer hidden states of the corresponding subwords, and denote the representation sequence as
\begin{equation}
    \mathbf{H}
    =
    [\mathbf{h}_1,\ldots,\mathbf{h}_T]
    \in\mathbb{R}^{T\times d}.
\end{equation}
All token-level evidence scores are computed on these word-level representations.

\section{Methodology}
\label{sec:method}

In this section, we propose \ourmethod, a dual-evidence framework with adaptive fusion and aggregation for token-level text anomaly detection. As illustrated in Figure~\ref{fig:method_overview}, to establish a more comprehensive score estimation process for each token, we first propose a dual-evidence score construction module that constructs a form-structural score and a semantic score, respectively (Sec.~\ref{subsec:score_construction}). The scores are then fused with an adaptive weighting scheme (Sec.~\ref{subsec:token_fusion}) and aggregated into the final document score via a multivariate aggregation module (Sec.~\ref{subsec:document_aggregation}). Moreover, we provide details on the training and inference stages of \ourmethod under the one-class setting where anomaly labels are unavailable (Sec.~\ref{subsec:training_inference}).

\subsection{Dual-Evidence Score Construction}
\label{subsec:score_construction}

Existing representation-based methods typically score each token in a pretrained contextual embedding space~\cite{lee2025effective}. While this formulation is quite straightforward, it can obscure diverse types of token abnormality.
For example, a sentiment-bearing word in a review may be semantically suspicious despite having a normal surface form, whereas a corrupted string in a spam message may be visually abnormal yet still receive a plausible contextual representation.
To fill the gap, we establish the scoring process with two complementary views, i.e., form-structural view and semantic view, to capture the surface-form extrinsic anomaly and the intrinsic semantic anomaly, respectively. Consequently, two distinctive token scores, $s_i^{\form}$ and $s_i^{\sem}$ are computed in parallel within the dual-view evidence modeling process, providing a more comprehensive perspective for token-level anomaly localization.

\noindent\textbf{Form-Structural Score.}
Form-structural anomalies are characterized by visible irregularities in surface form or local structure, e.g., character corruption, repeated words, and unusual punctuation. 
Usually, they are easier to capture through token-level analysis rather than through highly abstracted contextual embeddings as clues. 
In this case, relying solely on representations may smooth out discriminative surface-level patterns and weaken the detection of such anomalies. Hence, instead of using representations, we build a statistical token analysis module to extract compact token descriptors and derive the form-structural score.

To identify form-structural anomalies, we first construct a descriptor that extracts key features for anomaly pattern identification. The descriptor aims to measure whether a token is unusual in its string form or its local context.
Specifically, for each token $w_i$, \ourmethod computes a 16-dimensional descriptor
\begin{equation}
    \mathbf{d}_i =
    [
    \mathbf{d}^{\mathrm{surf}}_i;
    \mathbf{d}^{\mathrm{ctx}}_i
    ],
\end{equation}
where $\mathbf{d}^{\mathrm{surf}}_i \in \mathbb{R}^{10}$ contains surface statistics computed directly from the token string, and $\mathbf{d}^{\mathrm{ctx}}_i \in \mathbb{R}^{6}$ contains embedding-based local context features computed from neighboring word embeddings.
Consequently, the \textit{surface part} $\mathbf{d}^{\mathrm{surf}}_i$ contains directly computed character-level statistics, such as length, digit ratios, letters, uppercase characters, and punctuation, repetition, uniqueness, and entropy, which capture string-level irregularities without relying on contextual embeddings. For instance, corrupted forms such as ``\texttt{recieve}'' or ``\texttt{gr8t}'' can be reflected in $\mathbf{d}^{\mathrm{surf}}_i$. 
Meanwhile, the \textit{contextual part} $\mathbf{d}^{\mathrm{ctx}}_i$ is derived from the embeddings of the two neighboring tokens on each side of $w_i$ on the basis of geometric cues such as token-context similarity, Euclidean distance, norm ratio, and left-right consistency. 
It does not rely solely on the token itself, but instead measures whether structural deviations exist within its local context.  
For example, repeated determiners in ``\texttt{the the result}'' may look normal individually but produce abnormal local geometry captured by $\mathbf{d}^{\mathrm{ctx}}_i$. 
Rather than serving as manual anomaly rules, these descriptors automatically measure how the surface form and local context structure of a token differ from normal training tokens. 

Using descriptors based on normal training tokens, we fit a Gaussian distribution to normal-token descriptors with mean $\boldsymbol{\mu}_d$ and covariance $\boldsymbol{\Sigma}_d$.
Then, the form-structural score is defined as the Mahalanobis distance:
\begin{equation}
\begin{aligned}
    s_i^{\form}
    =
    \sqrt{
    (\mathbf{d}_i-\boldsymbol{\mu}_d)^{\top}
    \boldsymbol{\Sigma}_d^{-1}
    (\mathbf{d}_i-\boldsymbol{\mu}_d)
    },
\end{aligned}
\label{eq:s_i_form}
\end{equation}
where a larger score indicates that the token is more unusual in surface form or local structure compared with normal training tokens, providing evidence for identifying form-structural anomalies. 
By operating on compact token descriptors rather than highly abstracted contextual embeddings, this scoring scheme preserves surface-level abnormal signals that may otherwise be smoothed out in PLM-derived representations.

\noindent\textbf{Semantic Score.}
While the form-structural score captures abnormal strings and local structural deviations, it cannot fully detect tokens that are surface-form normal but semantically incompatible with their surrounding context. To capture such semantic deviations, a straightforward solution is to employ an encoder--decoder architecture to compute reconstruction errors for identifying abnormal tokens, i.e., a larger reconstruction error of representation can indicate semantic anomalies. However, since the architecture is usually trained on a large collection of normal documents, sometimes it becomes overly expressive and partially reconstructs abnormal inputs, thereby reducing the discriminative power of reconstruction error for semantic anomaly detection. 

To solve this problem, we introduce a self-distilled teacher--student discrepancy signal to strengthen the detections that are suppressed by direct reconstruction~\cite{tang2026advancing}. 
In particular, the semantic branch consists of an encoder $f_{\theta}$, a teacher decoder $g_{\phi}$, and a lightweight student decoder $g_{\psi}$:
\begin{equation}
\begin{aligned}
    \mathbf{Z} &= f_{\theta}(\mathbf{H}),\\
    \widehat{\mathbf{H}}^{T} &= g_{\phi}(\mathbf{Z}),\\
    \widehat{\mathbf{H}}^{S} &= g_{\psi}(\mathbf{Z}),
\end{aligned}
\end{equation}
where $f_\theta$ and $g_\phi$ are well-trained on normal documents by reconstructing the original contextual representations.
After this training, the encoder and teacher decoder are frozen, and a lightweight student decoder $g_{\psi}$ is optimized to reproduce the reconstruction behavior of the teacher decoder based on the same normal input data.
Since the student decoder is trained only to mimic the teacher decoder on normal data, abnormal tokens can lead to a larger teacher--student discrepancy, which can serve as an effective signal for identifying semantic anomalies.
For instance, a sentiment word such as ``\texttt{impressive}'' in a strongly negative review may still be partially reconstructed by the teacher, as the teacher decoder has sufficient prior knowledge to recover some less appropriate expressions.
In contrast, the student is likely to deviate from the teacher and keep generating negative outputs aligned with the context, due to its limited exposure to such unaligned context. 
As a result, the distinctive outputs between teacher and student decoders make the anomaly more distinguishable and can be translated into a teacher--student discrepancy signal for semantic anomaly detection. 
Specifically, the $i$-th row of $\widehat{\mathbf{H}}^{T}$ and $\widehat{\mathbf{H}}^{S}$, i.e., $\widehat{\mathbf{h}}^{T}_i$ and $\widehat{\mathbf{h}}^{S}_i$, denotes the reconstructed representation of token $w_i$ produced by the teacher decoder and the student decoder, respectively. 
Given two vectors $\mathbf{a}$ and $\mathbf{b}$, the distance operator is defined as
\begin{equation}
\begin{aligned}
    d(\mathbf{a},\mathbf{b})
    =
    \|\mathbf{a}-\mathbf{b}\|_2^2
    +
    \lambda\bigl(1-\cos(\mathbf{a},\mathbf{b})\bigr).
\end{aligned}
\end{equation}
Then, the semantic anomaly score can be computed as
\begin{equation}
\begin{aligned}
    s_i^{\sem}
    =
    d(\widehat{\mathbf{h}}^{T}_i,\mathbf{h}_i)
    +
    d(\widehat{\mathbf{h}}^{S}_i,\widehat{\mathbf{h}}^{T}_i),
\end{aligned}
\label{eq:s_i_sem}
\end{equation}
where the first term is the teacher reconstruction error, directly measuring whether the original token embedding can be recovered from normal contextual patterns, and the second term is the teacher--student discrepancy, indicating how much the student output deviates from the teacher output. 
For normal tokens, both terms are expected to remain small, whereas semantic anomalies can increase the score through reconstruction failure, teacher--student mismatch, or both. 
This design provides a more robust semantic anomaly signal than relying on reconstruction error alone, as the discrepancy term helps capture anomalies that lie close to normal documents in the representation space. 
Moreover, combining the two terms allows the semantic score to capture both absolute reconstruction difficulty and relative disagreement between the teacher and student decoders.

\subsection{Calibration-based Token-Level Score Fusion}
\label{subsec:token_fusion}

\begin{algorithm}[t]
\caption{Training and inference procedure of \ourmethod{}}
\label{alg:desa_training}
\begin{algorithmic}[1]

\REQUIRE Normal corpus $\mathcal{D}_n$; pretrained language model $\mathrm{PLM}$; pseudo-anomaly generator $\mathcal{G}$
\ENSURE Token scores $\{s_i\}_{i=1}^{T}$ and document score $S(x)$

\STATE Extract and cache representations $\{\mathbf{H}(x):x\in\mathcal{D}_n\}$ using $\mathrm{PLM}$.
\STATE Initialize semantic modules $f_\theta,g_\phi,g_\psi$ and scoring modules $q_\omega,r_\eta$.

\STATE \textbf{// Learn normal references}
\STATE Train $f_\theta$ and $g_\phi$ on normal representations, then train $g_\psi$ to mimic $g_\phi$.
\STATE Fit structural statistics $(\boldsymbol{\mu}_d,\boldsymbol{\Sigma}_d)$ on normal token descriptors.
\STATE Construct semantic and structural references $\mathcal{R}^{\sem}$ and $\mathcal{R}^{\form}$.

\STATE \textbf{// Pseudo-anomaly training}
\STATE Generate pseudo anomalies $\widetilde{\mathcal{D}}=\mathcal{G}(\mathcal{D}_n)$.
\FORALL{$(\widetilde{x},\mathbf{y}^{tok},y^{doc})\in\widetilde{\mathcal{D}}$}
    \STATE Compute raw scores $\{s_i^{\sem},s_i^{\form}\}_{i=1}^{T}$.
    \STATE Calibrate scores using $\mathcal{R}^{\sem}$ and $\mathcal{R}^{\form}$.
    \STATE Fuse calibrated evidence with $q_\omega$ to obtain $\{s_i\}_{i=1}^{T}$.
    \STATE Compute candidate document scores $\{S_m(\widetilde{x})\}_{m\in\mathcal{M}}$.
    \STATE Predict aggregation weights with $r_\eta$ and obtain $S(\widetilde{x})$.
    \STATE Update $q_\omega$ and $r_\eta$ using token- and document-level labels.
\ENDFOR

\STATE \textbf{// Inference}
\FORALL{test document $x$}
    \STATE Compute $\{s_i^{\sem},s_i^{\form}\}_{i=1}^{T}$.
    \STATE Calibrate and fuse evidence to obtain $\{s_i\}_{i=1}^{T}$.
    \STATE Compute $\{S_m(x)\}_{m\in\mathcal{M}}$ and aggregation weights using $r_\eta$.
    \STATE Obtain document score $S(x)$.
    \STATE \textbf{return} $\{s_i\}_{i=1}^{T}$ and $S(x)$.
\ENDFOR

\end{algorithmic}
\end{algorithm}

After obtaining the form-structural score $s_i^{\form}$ and the semantic score $s_i^{\sem}$ for each token, we need to further combine them into a unified token-level anomaly score. Nevertheless, since these two scores are produced by different mechanisms and may exhibit different numerical scales, a proper calibration and fusion scheme is required to convert them into final token-level anomaly scores. 
To this end, we design a \textit{calibration module} that aligns the two scores onto a comparable evidence scale, and then use an \textit{adaptive fusion module} to learn how much each evidence view should contribute. 

\noindent\textbf{Tail-Evidence Calibration.}
Due to differences in terms of the interpretation and scale of the two raw scores, direct summation may cause the fused score to be dominated by the score with a larger numerical range; \ourmethod therefore converts each score into a tail-evidence value that reflects its rarity among normal training tokens. Specifically, for each view $v\in\{\sem,\form\}$, $\mathcal{R}^{v}$ is defined as the set of normal training scores under view $v$. 
For each original test score $s_i^v$, we compute its empirical upper-tail probability as
\begin{equation}
\begin{aligned}
    p_i^v
    &=
    \frac{1+N_i^v}{1+|\mathcal{R}^{v}|},
\end{aligned}
\end{equation}
where $N_i^v=\left|\{r\in\mathcal{R}^{v}:r\geq s_i^v\}\right|$ denotes the number of scores in $\mathcal{R}^{v}$ that are greater than or equal to the test score $s_i^v$. This upper-tail probability $p_i^v$ measures the empirical probability that a normal training score is no smaller than $s_i^v$ under view $v$, thus converting the raw score into a distribution-aware rarity measure. It tells how rare the test score is among normal training scores, i.e., the smaller $p_i^v$ is, the rarer the token is under the normal reference distribution. 

We further transform this probability into calibrated tail evidence using the negative logarithm,
\begin{equation}\label{eq:calibrate}
    e_i^v=-\log(p_i^v+\epsilon),
\end{equation}
where larger $e_i^v$ corresponds to stronger anomaly evidence, and $\epsilon$ is used for numerical stability. This calibration makes the two raw scores comparable in scale by expressing both as tail evidence. Once the two views are calibrated in this way, the remaining question is how much each view should contribute to the final token score.

\begin{table*}[t]
\centering
\small
\setlength{\tabcolsep}{3pt}
\caption{Performance comparison. The best results are \textbf{bolded} and the second-best results are \underline{underlined}.}
\label{tab:main}
\begin{tabularx}{\textwidth}{l *{8}{C}}
\toprule
& \multicolumn{2}{c}{\textbf{Spam}} 
& \multicolumn{2}{c}{\textbf{Review}} 
& \multicolumn{2}{c}{\textbf{Grammar}} 
& \multicolumn{2}{c}{\textbf{Average}} \\
\cmidrule(lr){2-3}\cmidrule(lr){4-5}\cmidrule(lr){6-7}\cmidrule(lr){8-9}
\textbf{Method} 
& AUROC & AUPRC 
& AUROC & AUPRC 
& AUROC & AUPRC 
& AUROC & AUPRC \\
\midrule

\rowcolor[HTML]{F5F5F5} 
\multicolumn{9}{l}{\textit{Token-level}} \\
LOF        & 0.5315 & 0.0098 & 0.8005 & 0.0569 & 0.5641 & 0.0339 & 0.6320 & 0.0335 \\
iForest    & 0.8157 & \underline{0.0341} & 0.7052 & 0.0768 & 0.5024 & 0.0303 & 0.6744 & 0.0471 \\
ECOD       & \underline{0.8186} & 0.0289 & 0.7035 & \underline{0.0872} & 0.4457 & 0.0268 & 0.6559 & \underline{0.0476} \\
DeepSVDD   & 0.7366 & 0.0250 & 0.6554 & 0.0491 & 0.4798 & 0.0312 & 0.6239 & 0.0351 \\
AE         & 0.3851 & 0.0074 & 0.6665 & 0.0210 & 0.6190 & 0.0386 & 0.5569 & 0.0223 \\
LUNAR      & 0.6734 & 0.0139 & 0.8254 & 0.0516 & 0.6053 & 0.0376 & 0.7014 & 0.0344 \\
TokenCore  & 0.6792 & 0.0141 & \underline{0.8271} & 0.0530 & \underline{0.6371} & \underline{0.0407} & \underline{0.7145} & 0.0359 \\
\ourmethod{} 
& \textbf{0.9923} & \textbf{0.4580} 
& \textbf{0.8909} & \textbf{0.4682} 
& \textbf{0.8738} & \textbf{0.1155} 
& \textbf{0.9190} & \textbf{0.3472} \\
\midrule

\rowcolor[HTML]{F5F5F5} 
\multicolumn{9}{l}{\textit{Document-level}} \\
LOF        & \underline{0.5948} & \underline{0.2276} & 0.9082 & 0.4107 & 0.5236 & 0.1933 & 0.6755 & 0.2772 \\
iForest    & 0.4598 & 0.1551 & 0.8951 & 0.3987 & 0.5994 & 0.2342 & 0.6514 & 0.2627 \\
ECOD       & 0.4603 & 0.1552 & 0.8887 & 0.3791 & 0.5739 & 0.2178 & 0.6410 & 0.2507 \\
DeepSVDD   & 0.4814 & 0.1633 & 0.8572 & 0.3371 & 0.5548 & 0.2294 & 0.6311 & 0.2433 \\
AE         & 0.4576 & 0.1566 & 0.9075 & 0.5247 & 0.6403 & 0.2547 & 0.6685 & 0.3120 \\
LUNAR      & 0.5814 & 0.2051 & 0.9587 & 0.7473 & \underline{0.6578} & \underline{0.2628} & 0.7326 & 0.4051 \\
TokenCore  & 0.5859 & 0.2072 & \underline{0.9594} & \underline{0.7495} & 0.6553 & 0.2624 & \underline{0.7335} & \underline{0.4064} \\
\ourmethod{} 
& \textbf{0.9067} & \textbf{0.5236} 
& \textbf{0.9782} & \textbf{0.8886} 
& \textbf{0.7311} & \textbf{0.3788} 
& \textbf{0.8720} & \textbf{0.5970} \\
\bottomrule
\end{tabularx}
\end{table*}

\noindent\textbf{Adaptive Fusion with Pseudo Anomalies.}
Although calibrated evidence makes the two views comparable, it does not determine which view should dominate for a particular token.
Different anomaly sources require different evidence: semantic anomalies should rely more on semantic evidence, while spelling, punctuation, or character-level errors should rely more on form-structural evidence.
To determine the token-specific contribution of each view, we introduce a lightweight adaptive fusion module that dynamically weights the two evidence sources and derives the token-level anomaly score.

To be specific, the fusion network predicts a semantic gate $\alpha_i$ and an agreement gate $\beta_i$:
\begin{equation}
    \alpha_i,\beta_i=\sigma(q_{\omega}(\mathbf{u}_i)),
\end{equation}
where $\mathbf{u}_i$ contains compact evidence features such as log evidence, evidence difference, cross-view agreement, and maximum evidence.
The fused token score is
\begin{equation}\label{eq:fuse}
\begin{aligned}
    s_i
    &=
    \alpha_i e_i^{\sem}
    +(1-\alpha_i)e_i^{\form}
    \\
    &\quad+
    \beta_i\sqrt{e_i^{\sem}e_i^{\form}},
\end{aligned}
\end{equation}
where the first two terms adaptively balance semantic and form-structural evidence, while the last term gives additional weight to tokens that are suspicious under both views.

Since real anomaly labels are unavailable during model training, it is not feasible to leverage ground-truth supervision to learn such view-specific fusion behavior. To address this issue, we learn this fusion behavior from pseudo anomalies generated from normal data.
Specifically, we create pseudo anomalies of three perturbation types: string mutation for form-structural anomalies, representation replacement for semantic anomalies, and applying both perturbations for mixed anomalies~\cite{qiao2024generative}.
For document-level training, we additionally generate diffuse variants in which several token positions are perturbed simultaneously, exposing the document aggregator to broader anomaly patterns.

\subsection{Document-Level Multivariate Score Aggregation}
\label{subsec:document_aggregation}

After obtaining the token-level anomaly score, the next step is to aggregate them to derive the document-level anomaly score. The main challenge here is that usually a document contains only a few anomalous tokens; therefore, a single aggregation strategy can smooth out the differences between documents, making anomalous documents appear indistinguishable from normal ones. For example, mean pooling can dilute rare abnormal tokens, while max pooling can mistake an isolated noisy peak for an anomaly. 

To deal with this problem, in \ourmethod, we employ multivariate aggregation to obtain the document-level score. Given the fused token scores $\mathbf{s}=[s_1,\ldots,s_T]$, we compute four candidate document scores using different pooling strategies including \textit{maximum pooling}, \textit{top-$k$ pooling}, \textit{log-sum-exp pooling}, and \textit{adaptive pooling}, and denote these scores as $S_{\max}(x)$, $S_{\mathrm{top}k}(x)$, $S_{\mathrm{lse}}(x)$, and $S_{\mathrm{adp}}(x)$~\cite{mcfee2018adaptive}, respectively. These four scores are then jointly modeled to produce the final document score. Compared with using a single pooling rule, this multivariate design captures complementary aspects of the token-score distribution, including the strongest abnormal evidence, the concentration of high-scoring tokens, and the overall distributional tendency, thereby providing a more robust and flexible document-level abnormality measure.

Then, to adaptively combine these candidate scores, we construct a 9-dimensional score-shape descriptor $\mathbf{a}(x)$ that captures the document-specific token-score pattern and serves as the basis for estimating the importance of the candidate scores. 
The descriptor contains the four candidate scores themselves, together with five auxiliary statistics that summarize the token-score distribution: the mean score, peakness, support, a geometry gate, and the log document length. 
Here, the four candidate scores play a dual role. First, they are included in $\mathbf{a}(x)$ to help the MLP infer which pooling strategy should be trusted more for the current document. Second, they are also the components to be combined by the predicted mixture weights to produce the final document-level score. To implement this adaptive weighting mechanism, a small MLP $r_{\eta}$ is employed, 
\begin{equation}
    \boldsymbol{\pi}(x)=\mathrm{softmax}(r_{\eta}(\mathbf{a}(x))),
\end{equation}
which takes $\mathbf{a}(x)$ as input and outputs four document-specific mixture weights, and each element in $\boldsymbol{\pi}(x)$ corresponds to one candidate aggregation score. 

Let $\mathcal{M}=\{\max,\mathrm{top}k,\mathrm{lse},\mathrm{adp}\}$ denote the set of candidate document scoring strategies, the final document anomaly score is computed as
\begin{equation}\label{eq:document_score}
    S(x)=
    \sum_{m\in\mathcal{M}}
    \pi_m(x)S_m(x),
\end{equation}
where $S_m(x)$ is the document score produced by strategy $m\in\mathcal{M}$ and $\pi_m(x)$ is the predicted weight assigned to strategy $m$. 
Powered by this aggregation module, \ourmethod
learns document-specific aggregation weights from the token-level anomaly score distribution. This enables the model to adapt to different anomaly patterns, such as isolated sharp anomalies or more diffuse abnormal evidence, leading to a more flexible and reliable document-level anomaly score.

\begin{table*}[t]
\centering
\small
\setlength{\tabcolsep}{3pt}
\caption{Component ablation. \textit{Top:} token-level performance of individual evidence views and their learned fusion. \textit{Bottom:} document-level performance of different aggregation rules applied to the same fused token scores.}
\label{tab:ablation}
\begin{tabularx}{\textwidth}{l *{8}{C}}
\toprule
& \multicolumn{2}{c}{\textbf{Spam}} 
& \multicolumn{2}{c}{\textbf{Review}} 
& \multicolumn{2}{c}{\textbf{Grammar}}
& \multicolumn{2}{c}{\textbf{Average}} \\
\cmidrule(lr){2-3}\cmidrule(lr){4-5}\cmidrule(lr){6-7}\cmidrule(lr){8-9}
\textbf{Variant} & AUROC & AUPRC & AUROC & AUPRC & AUROC & AUPRC & AUROC & AUPRC \\
\midrule

\rowcolor[HTML]{F5F5F5} \multicolumn{9}{l}{\textit{Token-level evidence view}} \\
Semantic only        & \underline{0.9860} & \underline{0.2601} & \textbf{0.9082} & 0.2484 & 0.5745 & 0.0337 & 0.8229 & 0.1807 \\
Form-structural only & 0.9498 & 0.1298 & 0.7894 & \underline{0.3882} & \textbf{0.9437} & \textbf{0.2506} & \underline{0.8943} & \underline{0.2562} \\
\ourmethod{} & \textbf{0.9923} & \textbf{0.4580} & \underline{0.8909} & \textbf{0.4682} & \underline{0.8738} & \underline{0.1155} & \textbf{0.9190} & \textbf{0.3472} \\
\midrule

\rowcolor[HTML]{F5F5F5} \multicolumn{9}{l}{\textit{Document-level aggregation}} \\
Mean                 & 0.7972 & 0.3632 & 0.9332 & 0.4316 & \underline{0.7583} & 0.3348 & 0.8296 & 0.3765 \\
Max                  & \underline{0.8946} & \underline{0.4841} & 0.9706 & \underline{0.8796} & 0.7064 & 0.2976 & 0.8572 & \underline{0.5538} \\
Top-$k$              & 0.8705 & 0.4557 & \underline{0.9726} & 0.8157 & \textbf{0.7689} & \underline{0.3539} & \underline{0.8707} & 0.5418 \\
\ourmethod{}         & \textbf{0.9067} & \textbf{0.5236} & \textbf{0.9782} & \textbf{0.8886} & 0.7311 & \textbf{0.3788} & \textbf{0.8720} & \textbf{0.5970} \\
\bottomrule
\end{tabularx}
\end{table*}

\subsection{Training and Inference}
\label{subsec:training_inference}
For training and inference (see Algorithm~\ref{alg:desa_training}), we follow the standard one-class anomaly detection setting, where only normal documents are available during training and real anomaly labels are only available for evaluation.

During \textbf{training}, \ourmethod separates the learning of normal patterns from the learning of fusion and aggregation modules on synthetic anomalies.
Specifically, in the \textit{first stage}, where only normal documents are used, the teacher model learns to reconstruct contextual representations, while the student model learns to mimic the teacher on normal data. Meanwhile, the form-structural descriptor distribution is fitted to normal tokens.
These learned components are then fixed and used to compute semantic and form-structural evidence. 
In the \textit{second stage}, pseudo anomalies are generated from normal documents by perturbing either a single token or a small subset of tokens, exposing the document scorer to both sharp peaks and more diffuse score patterns.
Then, these pseudo-anomalous documents are used to train the token fusion module and the document scorer, enabling the model to learn how to fuse and aggregate the evidence produced by the fixed modules.

During \textbf{inference}, given a test document, \ourmethod first computes form-structural and semantic scores for each token using the fixed references, according to Eq.~\eqref{eq:s_i_form} and Eq.~\eqref{eq:s_i_sem}, respectively. Next, these two scores are calibrated into tail evidence (Eq.~\eqref{eq:calibrate}) and fused into token-level anomaly scores (Eq.~\eqref{eq:fuse}).
Then, token-level scores are passed to the document scorer to obtain the document-level anomaly score $S(x)$ following Eq.~\eqref{eq:document_score}.
Thus, \ourmethod performs dual evidence fusion and aggregation in a unified inference pipeline, allowing complementary token-level evidence to be effectively integrated and translated into reliable document-level anomaly scores. 

\subsection{Computational Complexity}
We analyze the computational complexity of \ourmethod{} for a document with $T$ word-level tokens and embedding dimension $d$. 
The form-structural branch computes surface descriptors and local context descriptors with complexity $O(TW+Tkd)$, where $W$ is the average token length and $k$ is the local window size; fitting the Gaussian reference on $N$ normal tokens costs $O(Nd_d+Nd_d^2)$ with fixed descriptor dimension $d_d=16$. 
The semantic branch uses a fixed-depth Transformer encoder and teacher decoder, together with a lightweight MLP student decoder, resulting in per-document semantic scoring cost $O(T^2d+Td^2)$. 
Tail-evidence calibration with sorted normal-score references costs $O(T\log N)$, adaptive fusion costs $O(T)$, and document aggregation costs at most $O(T\log k)$. 
Therefore, excluding the pretrained language model forward pass, the overall inference complexity is $O\bigl(T^2d + Td^2 + T(W+kd+\log N+\log k)\bigr)$. 

\section{Experiments}
\label{sec:experiments}

\subsection{Experimental Setup}
\label{subsec:exp_setup}

\begin{figure*}[t]
    \centering
    \subfigure[Accuracy--efficiency Pareto comparison on Grammar.]{
        \begin{minipage}[b]{0.3\textwidth}
            \includegraphics[width=1\textwidth]{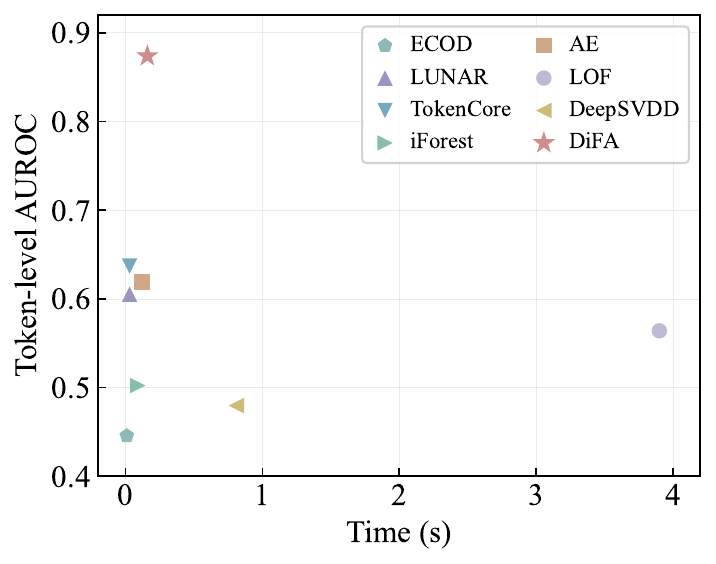}
        \end{minipage}
        \label{fig:efficiency}
    }
    \subfigure[Training-reference contamination on Spam.]{
        \begin{minipage}[b]{0.3\textwidth}
        \includegraphics[width=1\textwidth]{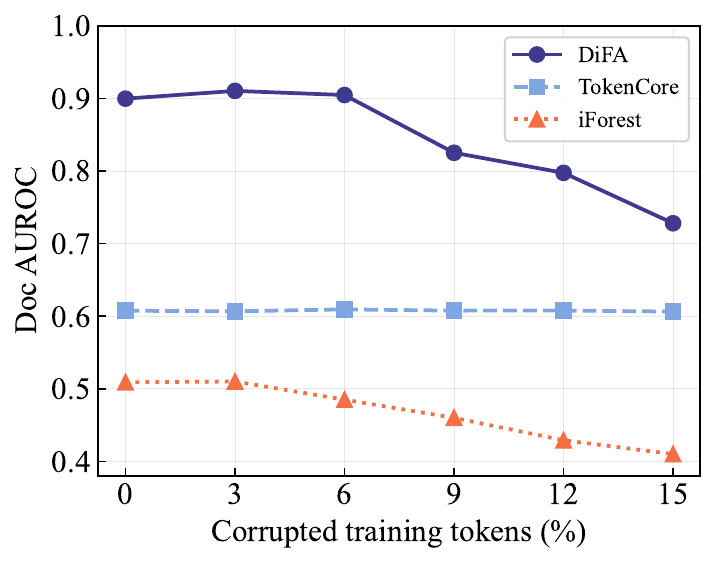}
        \end{minipage}
        \label{fig:analysis_contamination}
    }
    \subfigure[Controlled anomaly sparsity on Spam.]{
        \begin{minipage}[b]{0.3\textwidth}
        \includegraphics[width=1\textwidth]{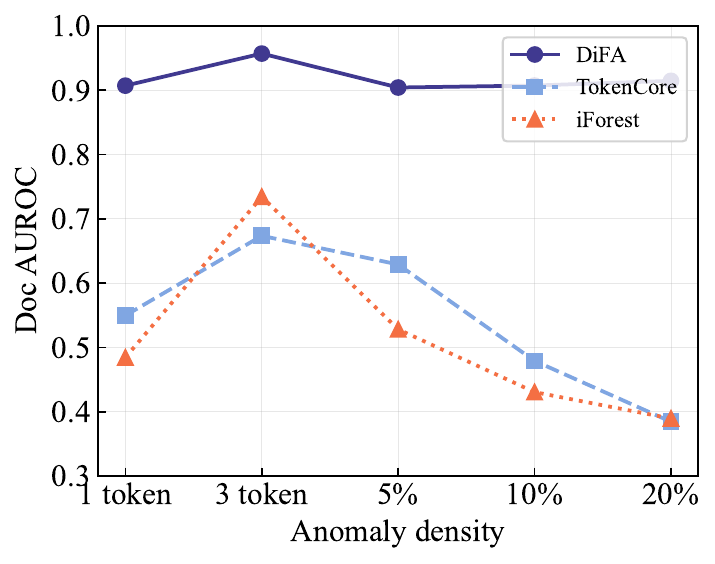}
        \end{minipage}
        \label{fig:analysis_sparsity}
    }
    
    \caption{
    Analysis results.
    (a) Accuracy--efficiency Pareto comparison on Grammar.
    (b) Document-level AUROC under training-reference contamination on Spam.
    (c) Document-level AUROC under controlled anomaly sparsity on Spam.
    }
    \label{fig:analysis_results}
\end{figure*}

\begin{figure}[t]
\centering
\includegraphics[width=\columnwidth]{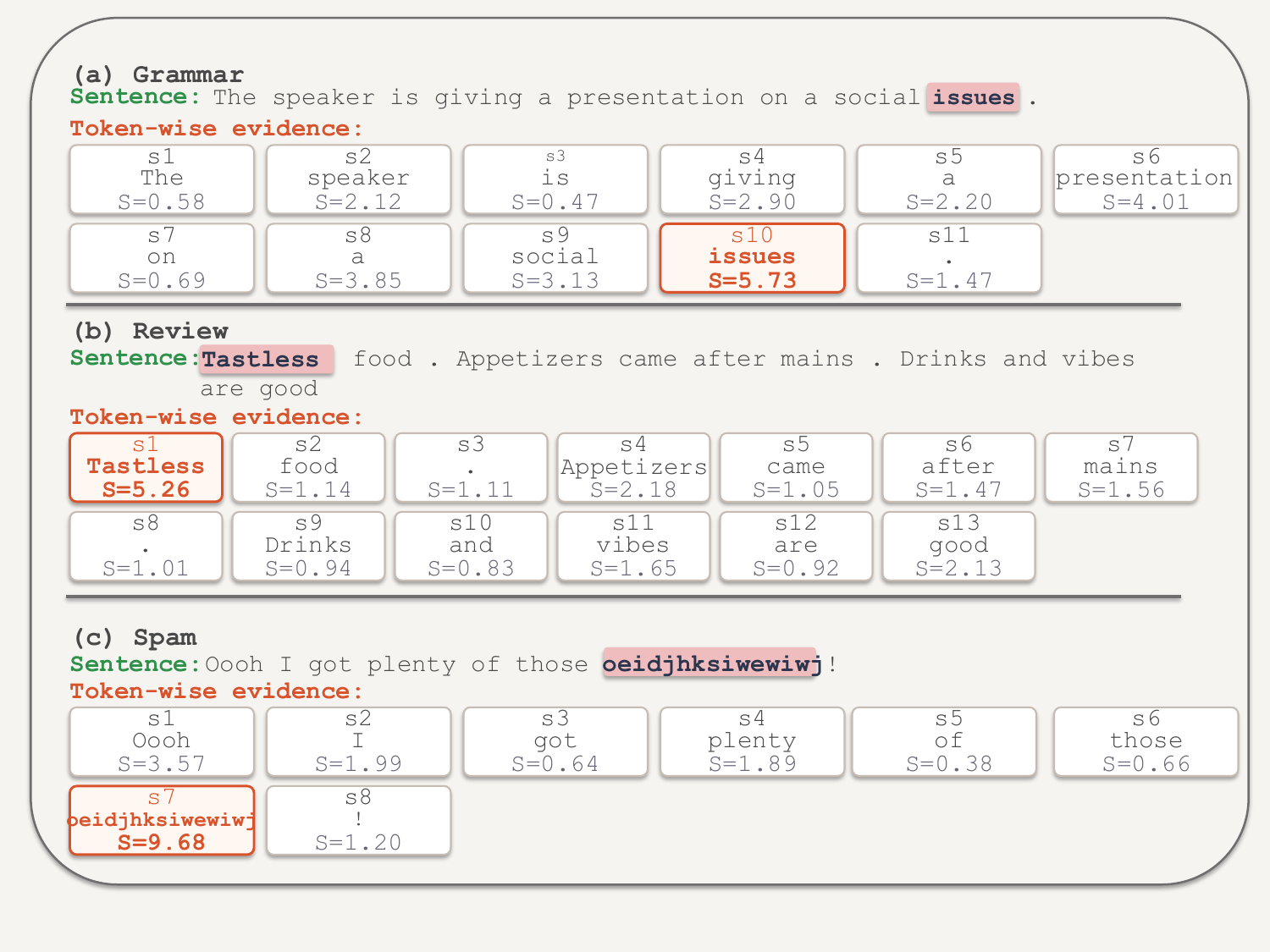}
\caption{Qualitative localization results.}
\label{fig:e5}
\end{figure}

\subsubsection{Datasets and Models}
To ensure a fair and rigorous comparison, we evaluate \ourmethod{} on three token-level text anomaly detection datasets: Grammar, Review, and Spam. They feature distinct anomaly types: Grammar covering form-structural errors, Review targeting semantic anomalies in customer reviews, and Spam containing surface-form corruptions or abnormal textual patterns. Each dataset provides token-level annotations for localization as well as document-level labels for detection~\cite{cao2026towards}. 

\subsubsection{Baselines}
We compare \ourmethod{} with representative anomaly detection methods under the same benchmark setting~\cite{cao2026towards}. The baselines include LOF, Isolation Forest, ECOD, DeepSVDD, AutoEncoder, LUNAR, and TokenCore, covering density-based, isolation-based, statistical, reconstruction-based, deep one-class, graph-based, and memorybank-based detection paradigms~\cite{breunig2000lof,li2022ecod,ruff2018deep,goodge2022lunar,cao2026towards}. All methods use the same BERT-base-uncased representations, so the comparison focuses on scoring, fusion, and aggregation mechanisms.

\subsubsection{Evaluation Protocol}
We evaluate performance at both token and document levels using AUROC and AUPRC.
Token-level evaluation measures whether anomalous words can be correctly localized, while document-level evaluation measures whether fused token-level evidence can be converted into reliable document-level decisions.
Model selection is based on validation performance, and all reported results are obtained on the held-out test set.

\subsection{Experimental Results}
\label{subsec:results}

\subsubsection{Main Results}
Table~\ref{tab:main} reports token-level and document-level anomaly detection results on Spam, Review, and Grammar. 
We summarize the main observations as follows. 
\ding{182}~\ourmethod{} achieves the best performance across all datasets and metrics, showing consistent advantages over representative anomaly detection baselines.
\ding{183}~The gains are particularly clear at the token level, where anomalous words are sparse and difficult to localize; the large AUPRC improvements on Spam indicate stronger token-level anomaly evidence. 
\ding{184}~At the document level, the strong results show that localized token evidence can be effectively transferred to document-level decisions.

\subsubsection{Component Ablation}
Table~\ref{tab:ablation} decomposes \ourmethod{} by evidence view and document aggregation strategy.
\ding{182}~Different datasets favor different evidence sources: form-structural evidence is most effective on Grammar, semantic evidence is stronger on Review, and both views contribute on Spam. 
This confirms that token-level anomalies are heterogeneous and cannot be reliably captured by a single evidence view.
\ding{183}~For document-level aggregation, fixed pooling rules show inconsistent behavior across datasets: mean pooling can dilute sparse anomalies, while peak-sensitive rules such as max and top-$k$ can be affected by isolated noise.
Overall, the model remains competitive, showing the benefit of adapting document scoring to each document's token-score distribution.

\subsubsection{Efficiency Analysis}
Figure~\ref{fig:efficiency} compares token-level AUROC and inference time on Grammar.
\ourmethod{} achieves the highest AUROC while remaining within the most efficient group of methods.
Although a few lightweight baselines are slightly faster, their detection performance is much lower.
These results show that \ourmethod{} lies on the Pareto frontier of accuracy and efficiency, achieving a large AUROC advantage without introducing prohibitive runtime overhead.

\subsubsection{Training-Reference Robustness}
Figure~\ref{fig:analysis_contamination} evaluates robustness when a fraction of training token representations in the Spam training set are perturbed with Gaussian noise.
\ourmethod{} maintains strong document-level AUROC under low and moderate corruption and degrades gradually as the corruption ratio increases. 
In contrast, TokenCore and iForest remain substantially lower across all corruption levels. 
These results show that \ourmethod{} remains stable under moderate corruption of the normal training references.

\subsubsection{Anomaly Sparsity Robustness}
Figure~\ref{fig:analysis_sparsity} evaluates document-level detection under different anomaly sparsity levels. 
\ourmethod{} maintains consistently high AUROC across different settings, whereas TokenCore and iForest degrade markedly as the anomaly distribution changes. 
This demonstrates that multivariate aggregation is important for document-level detection, especially when anomalous evidence is concentrated in only a few tokens or appears in irregular score patterns.

\subsubsection{Qualitative Localization}
Figure~\ref{fig:e5} visualizes fused token-level anomaly scores on representative examples from the three datasets.
In the Grammar example, \ourmethod{} assigns the highest score to the form-structural error ``\texttt{issues}'' in ``\texttt{a social issues}''. 
In the Review example, semantically inconsistent expressions such as ``\texttt{Tastless}'' receive higher scores than surrounding normal tokens. 
In the Spam example, the surface-form corruption ``\texttt{oeidjhksiwewiwj}'' receives the strongest anomaly score. 
These examples show that \ourmethod{} can localize different types of token anomalies while keeping most normal tokens suppressed.

\section{Conclusion}
\label{sec:conclusion}

In this paper, we proposed \ourmethod{}, a dual-evidence framework with adaptive fusion and aggregation for token-level text anomaly detection. 
\ourmethod{} models token abnormality from form-structural and semantic views, calibrates both evidence types into tail evidence, and fuses them into token-level anomaly scores. 
It further uses multivariate aggregation to convert localized token evidence into document-level decisions without a single fixed pooling rule. 
Experiments on three benchmark datasets show that \ourmethod{} improves both token-level localization and document-level detection, demonstrating the value of heterogeneous anomaly evidence and adaptive token-to-document aggregation.

\section*{Acknowledgment}
The work of Y. Liu was partially supported by the Australian Research Council (ARC) under Grant No. DE260101172.

\bibliographystyle{IEEEtran}
\bibliography{custom}

\clearpage

\end{document}